\documentclass[acmtog,nonacm]{acmart}

\setcopyright{none}
\renewcommand\footnotetextcopyrightpermission[1]{}
\authorsaddresses{}            
\usepackage{graphicx}
\usepackage{subcaption}
\usepackage{amsmath}
\usepackage{booktabs}
\usepackage{multirow}         
\usepackage{xspace}
\usepackage{pifont}           
\usepackage{animate}          

\newcommand{\eg}{e.g.\@\xspace}

\newcommand{\cmark}{\ding{51}}   

\title{MoCapAnything V2: End-to-End Motion Capture for Arbitrary Skeletons}

\author{Kehong Gong}
\authornote{Equal contribution.}
\affiliation{%
  \institution{Huawei Technologies Co., Ltd}
  \city{Singapore}
  \country{Singapore}
}
\email{[email protected]}

\author{Zhengyu Wen}
\authornotemark[1]
\affiliation{%
  \institution{Huawei Central Research Institute}
  \city{Singapore}
  \country{Singapore}
}
\email{[email protected]}

\author{Dao Thien Phong}
\authornotemark[1]
\affiliation{%
  \institution{Huawei Central Research Institute}
  \city{Singapore}
  \country{Singapore}
}
\email{[email protected]}

\author{Mingxi Xu}
\affiliation{%
  \institution{Huawei Central Research Institute}
  \city{Singapore}
  \country{Singapore}
}
\email{[email protected]}

\author{Weixia He}
\affiliation{%
  \institution{Huawei Central Research Institute}
  \city{Singapore}
  \country{Singapore}
}
\email{[email protected]}

\author{Qi Wang}
\affiliation{%
  \institution{Huawei Central Research Institute}
  \city{Singapore}
  \country{Singapore}
}
\email{[email protected]}

\author{Ning Zhang}
\affiliation{%
  \institution{Huawei Central Research Institute}
  \city{Singapore}
  \country{Singapore}
}
\email{[email protected]}

\author{Zhengyu Li}
\affiliation{%
  \institution{Huawei Central Research Institute}
  \city{Singapore}
  \country{Singapore}
}
\email{[email protected]}

\author{Guanli Hou}
\affiliation{%
  \institution{Huawei Central Research Institute}
  \city{Singapore}
  \country{Singapore}
}
\email{[email protected]}

\author{Dongze Lian}
\affiliation{%
  \institution{Huawei Central Research Institute}
  \city{Singapore}
  \country{Singapore}
}
\email{[email protected]}

\author{Xiaoyu He}
\affiliation{%
  \institution{Huawei Central Research Institute}
  \city{Singapore}
  \country{Singapore}
}
\email{[email protected]}

\author{Mingyuan Zhang}
\authornote{Corresponding author.}
\affiliation{%
  \institution{Huawei Central Research Institute}
  \city{Singapore}
  \country{Singapore}
}
\email{[email protected]}

\author{Hanwang Zhang}
\affiliation{%
  \institution{Huawei Central Research Institute}
  \city{Singapore}
  \country{Singapore}
}
\email{[email protected]}

\renewcommand{\shortauthors}{Gong, Wen and Dao et al.}

\begin{abstract}
Recent methods for arbitrary-skeleton motion capture from monocular video~\cite{mocapanythingv1} follow a factorized pipeline, where a Video-to-Pose network predicts joint positions and an analytical inverse-kinematics (IK) stage recovers joint rotations. While effective, this design is inherently limited, since joint positions do not fully determine rotations and leave degrees of freedom such as bone-axis twist ambiguous, and meanwhile the non-differentiable IK stage prevents the system from adapting to noisy predictions or optimizing for the final animation objective. In this work, we instead present the first fully end-to-end framework, where both Video-to-Pose and Pose-to-Rotation are learnable and jointly optimized. Crucially, we observe that the ambiguity in pose-to-rotation mapping arises from missing coordinate-system information, as the same joint positions can correspond to different rotations under different rest poses and local axis conventions, and to resolve this we introduce a reference pose--rotation pair from the target asset, which together with the rest pose not only anchors the mapping but also defines the underlying rotation coordinate system. This turns rotation prediction into a well-constrained conditional problem and enables effective learning. In addition, our model predicts joint positions directly from video without relying on mesh intermediates, which improves both robustness and efficiency, and both stages share a skeleton-aware Global-Local Graph-guided Multi-Head Attention (GL-GMHA) module for joint-level local reasoning and global coordination. Experiments on Truebones Zoo and Objaverse show that our method reduces rotation error from $\sim$17$^{\circ}$ to $\sim$10$^{\circ}$, and to 6.54$^{\circ}$ on unseen skeletons, while achieving $\sim$20$\times$ faster inference than mesh-based pipelines. Code, data, and an interactive Hugging Face demo will be available at \url{https://animotionlab.github.io/MoCapAnythingV2/}.
\end{abstract}



\begin{teaserfigure}
  \centering
  \includegraphics[width=\textwidth]{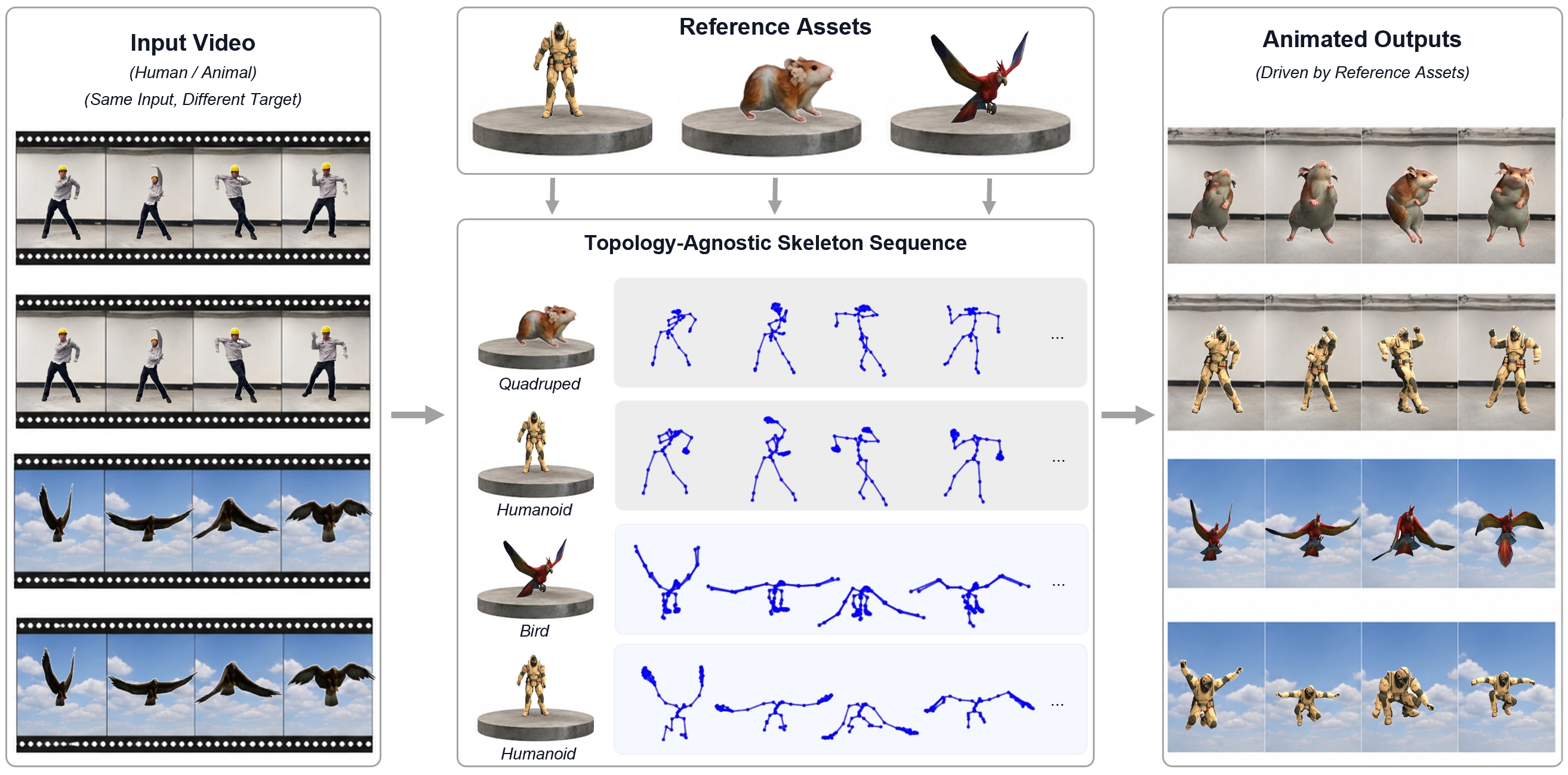}
  \caption{Overview of MoCapAnything V2. Given an input video of a human or an animal, our method infers a topology-agnostic skeleton sequence across diverse skeleton topologies. Conditioned on a reference asset, the model predicts animation-ready rotations via an end-to-end framework, enabling the reference asset to perform the input motion.
}
  \Description{Pipeline overview of the proposed method.}
\end{teaserfigure}

\begin{document}
\maketitle

\renewcommand{\thefootnote}{}%
\footnotetext{\textcopyright{} 2026 Copyright held by the owner/author(s). Published under a Creative Commons Attribution 4.0 International License. The definitive version appeared in \emph{ACM Transactions on Graphics}, Vol.~45, No.~6, Article~254 (December 2026), \url{https://doi.org/10.1145/3842550}.}%
\renewcommand{\thefootnote}{\arabic{footnote}}

\section{Introduction}
\label{sec:intro}

Recovering 3D character animation from monocular video, commonly referred to as motion capture, is a long-standing problem at the intersection of computer vision and computer graphics. While recent deep learning approaches have achieved strong performance for human motion capture~\cite{kanazawa2018end, pavllo:videopose3d:2019, kocabas2020vibe, goel2023humans}, extending this capability to the \emph{arbitrary-skeleton} setting remains fundamentally challenging. In this setting, the goal is to recover motion directly in the parameter space of a target rigged asset given only a monocular video and a reference skeleton~\cite{mocapanythingv1}. The difficulty arises from the large variation in skeleton topology, joint count, and especially local coordinate conventions across different assets.

A central challenge lies in recovering joint rotations. Unlike joint positions, rotations are defined with respect to a skeleton-specific coordinate system determined by the rest pose and local axis conventions. As a result, the same joint positions can correspond to different rotations under different skeletons, making direct video-to-rotation prediction prone to overfitting and poor generalization.

To mitigate this issue, existing methods adopt a \emph{factorized} design. A learned Video-to-Pose (V$\to$P) network first predicts 3D joint positions, and an analytical inverse-kinematics (IK) solver then converts these positions into joint rotations. This design leverages the fact that joint positions are largely shared across skeletons performing the same motion, thereby improving generalization at the pose level. However, this factorization introduces a fundamental limitation. Since joint positions do not fully determine rotations, the pose-to-rotation mapping is inherently under-constrained, and analytical solvers cannot resolve degrees of freedom such as bone-axis twist, nor can they adapt to the noise distribution of predicted poses. Moreover, the non-differentiable IK stage prevents the two components from being jointly optimized, forcing the pose predictor to ignore the downstream rotation objective.

In this work, we present the first \emph{fully end-to-end} framework for arbitrary-skeleton motion capture from monocular video, in which both the Video-to-Pose and Pose-to-Rotation (P$\to$R) stages are learnable and jointly optimized. The key challenge lies in the ill-posed nature of the P$\to$R mapping: the same 3D joint positions can correspond to different rotation values under different rest poses and local coordinate frames, and therefore pose alone is insufficient to determine rotations, as it does not specify how the target skeleton defines its rotation coordinate system. 

To address this issue, we extend the reference signal. While prior work relies on static geometry (skeleton, mesh, and rendered views)~\cite{mocapanythingv1}, we additionally introduce a single \emph{reference pose--rotation pair} from the target asset, which is naturally available for any rigged skeleton. Together with the rest pose, this reference acts as an explicit coordinate-system anchor, turning the ill-posed P$\to$R mapping into a well-constrained conditional prediction problem that can be modeled by a neural decoder. 

This design enables the P$\to$R stage to be learned, and more importantly, unlocks end-to-end training of the entire pipeline. Gradients from the rotation objective can now flow back through the pose intermediate and into the visual encoder, so that the pose representation is no longer optimized solely for positional accuracy, but is adaptively reshaped into a representation that better supports rotation recovery.

In addition, we remove the mesh intermediate used in prior work~\cite{mocapanythingv1}. While mesh representations can be beneficial when accurate, predicted mesh sequences often introduce noise and lead to error accumulation. By directly predicting joint positions from video, our approach improves robustness and significantly reduces computational cost, achieving approximately $20\times$ faster inference.

To support both stages, we introduce a skeleton-aware attention mechanism, \emph{Global-Local Graph-guided Multi-Head Attention} (GL-GMHA), which alternates between local kinematic-chain reasoning and global cross-branch coordination. This design enables effective modeling of both structural constraints and global motion dependencies across diverse skeletons.

We evaluate our method on the Truebones Zoo dataset, which covers seen, rare, and unseen skeletons, as well as the Objaverse benchmark. Our approach significantly improves rotation accuracy across all settings, especially on unseen skeletons, while also providing substantial gains in efficiency.

Our main contributions are as follows:
\begin{itemize}
    \item \textbf{End-to-end learnable Video-to-Pose-to-Rotation framework.} We present the first fully end-to-end trainable pipeline for arbitrary-skeleton motion capture, enabling joint optimization of pose and rotation.

    \item \textbf{Reference-conditioned rotation modeling.} We introduce a reference pose--rotation pair that, together with the rest pose, defines the rotation coordinate system and resolves the ambiguity of pose-to-rotation mapping.

    \item \textbf{Efficient direct video-to-pose prediction.} We eliminate the mesh intermediate, improving robustness and achieving approximately $20\times$ faster inference.

    \item \textbf{Skeleton-aware attention mechanism.} We propose GL-GMHA, which integrates local kinematic reasoning with global skeleton coordination in a unified attention framework.
\end{itemize}

\begin{figure*}[t]
  \centering
  \includegraphics[width=\textwidth]{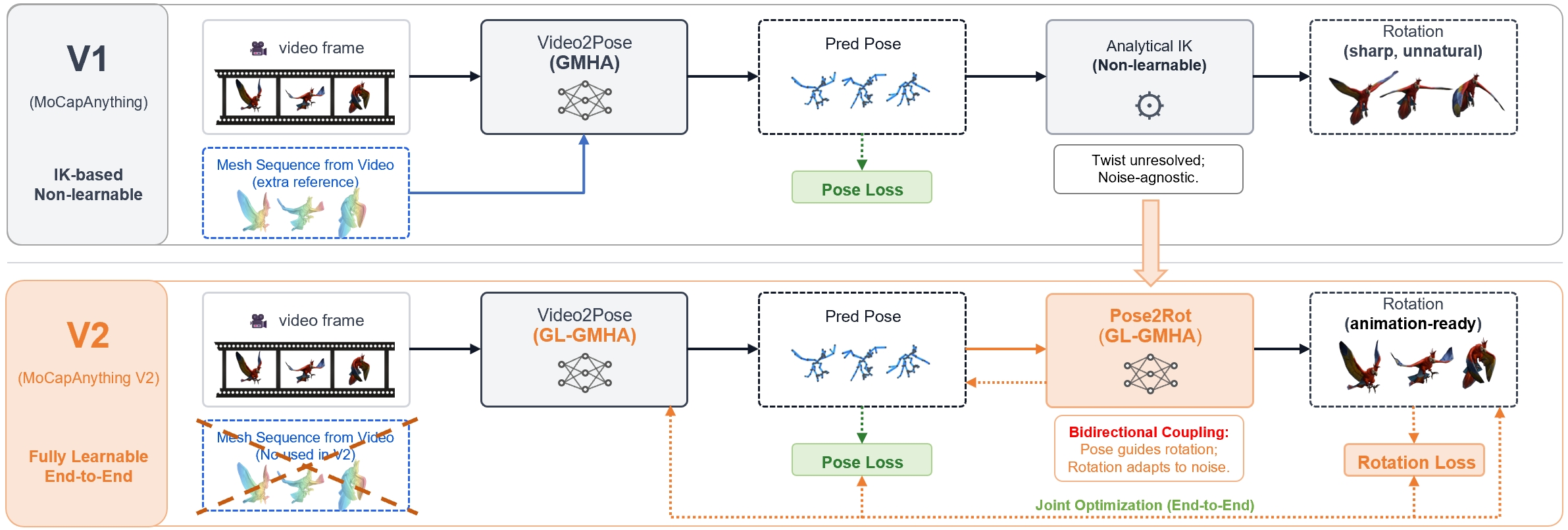}
  \caption{%
Comparison of MoCapAnything V1 and V2. Unlike V1, which depends on mesh-conditioned video-to-pose estimation and analytical inverse kinematics (IK) for rotation recovery, V2 eliminates mesh conditioning and introduces a fully learnable Pose2Rot module. The entire pipeline is optimized end-to-end, enabling bidirectional coupling between pose and rotation for improved robustness and animation-ready motion synthesis.
  }
  \Description{Side-by-side block-diagram comparison of two motion-capture pipelines. V1 (left) feeds video through a mesh predictor, then a pose predictor, then an analytical inverse-kinematics solver to produce rotations; the two learning stages are separated by a non-differentiable handoff. V2 (right) replaces the mesh stage and the analytical IK with a unified end-to-end neural pipeline in which Video-to-Pose and Pose-to-Rotation are jointly trained.}
  \label{fig:differencev1v2}
\end{figure*}

\section{Related Work}
\label{sec:related}

Recovering animation-ready motion from monocular video requires bridging the gap between observable visual cues (e.g., joint positions) and skeleton-specific motion parameters (e.g., rotations). Existing approaches address different aspects of this problem, but none fully resolve the ambiguity of mapping pose to rotation under varying skeleton definitions.

\paragraph{Pose estimation.}
Pose estimation methods localize anatomical keypoints from images or videos, ranging from heatmap-based architectures~\citep{sun2019deep} and DETR-style end-to-end models~\citep{xu2022vitpose, carion2020end, shi2022end, xiao2022querypose} to category-agnostic frameworks~\citep{xu2022pose, shi2023matching, hirschorn2023graph, ICLR2025_4f5aeaee} that generalize to unseen objects via support--query matching. However, these methods operate in 2D or in a fixed keypoint space and produce neither 3D motion trajectories nor skeleton-specific rotation parameters; in particular, they do not address how a joint configuration should be interpreted under different skeleton coordinate systems.

\paragraph{Motion capture.}
Monocular motion capture has been extensively studied for parametric human models~\citep{loper2015smpl, pavlakos2019expressive}, with feed-forward networks~\citep{kocabas2020vibe} and transformer-based variants regressing pose and shape parameters directly from video. Beyond humans, model-free reconstruction methods~\citep{kanazawa2018learning, wu2023magicpony, li2024learning} and video-based extensions~\citep{yang2022banmo, yang2021lasr} recover deformable surfaces, while category-specific parametric models such as SMAL~\citep{zuffi20173d} target animals. These approaches either operate in a fixed parameter space tied to predefined skeletons, or lack explicit skeletal parameterization, and therefore do not generalize to the arbitrary skeletons required for animation.
The closest line of work, MoCapAnything~\citep{mocapanythingv1}, conditions on a reference asset and adopts a factorized design: a learned Video-to-Pose network followed by an analytical IK solver. While pose as a shared intermediate improves cross-skeleton generalization, the mapping from positions to rotations is inherently under-constrained, since the same pose can correspond to different rotations under different rest poses and local coordinate systems, and the non-differentiable IK stage prevents end-to-end optimization. In contrast, we resolve the P$\to$R ambiguity with a reference pose--rotation pair that defines the underlying coordinate system, casting P$\to$R as a learnable conditional problem and training the entire V$\to$P$\to$R pipeline end-to-end.

\section{Method}
\label{sec:method}

\subsection{Problem Formulation}
\label{sec:problem}

Given an input video $\mathbf{V} = \{I_1, \dots, I_T\}$ and a target skeleton $\mathcal{S}$ defined by parent indices $\boldsymbol{\pi}$, bone offsets $\mathbf{o} \in \mathbb{R}^{J \times 3}$ (the rest pose), and per-joint semantic labels, we seek per-frame joint rotations $\mathbf{R} = \{\mathbf{r}_1, \dots, \mathbf{r}_T\}$ in 6D representation~\cite{zhou2019continuity}, $\mathbf{r}_t \in \mathbb{R}^{J \times 6}$, that drive $\mathcal{S}$ to perform the observed motion. $\mathcal{S}$ is assumed to be a tree-structured rigged skeleton with a single root; joint count $J$ and topology are otherwise arbitrary, padded up to $150$ to accommodate our largest training skeleton ($143$ joints).

To improve modeling stability and cross-skeleton generalization, we decompose the problem into two stages mediated by an explicit intermediate representation:
\begin{equation}
  \text{Video} \;\xrightarrow{\;\text{Stage 1}\;}\; \text{Pose} \;\xrightarrow{\;\text{Stage 2}\;}\; \text{Rotation},
  \label{eq:pipeline}
\end{equation}
where \emph{pose} $\mathbf{P} = \{\mathbf{p}_t\}$, $\mathbf{p}_t \in \mathbb{R}^{J \times 3}$, denotes root-relative 3D joint positions in the camera coordinate frame, and \emph{rotation} denotes local joint rotations in $\mathcal{S}$'s coordinate system.

\begin{figure*}[t]
  \centering
  \includegraphics[width=\textwidth]{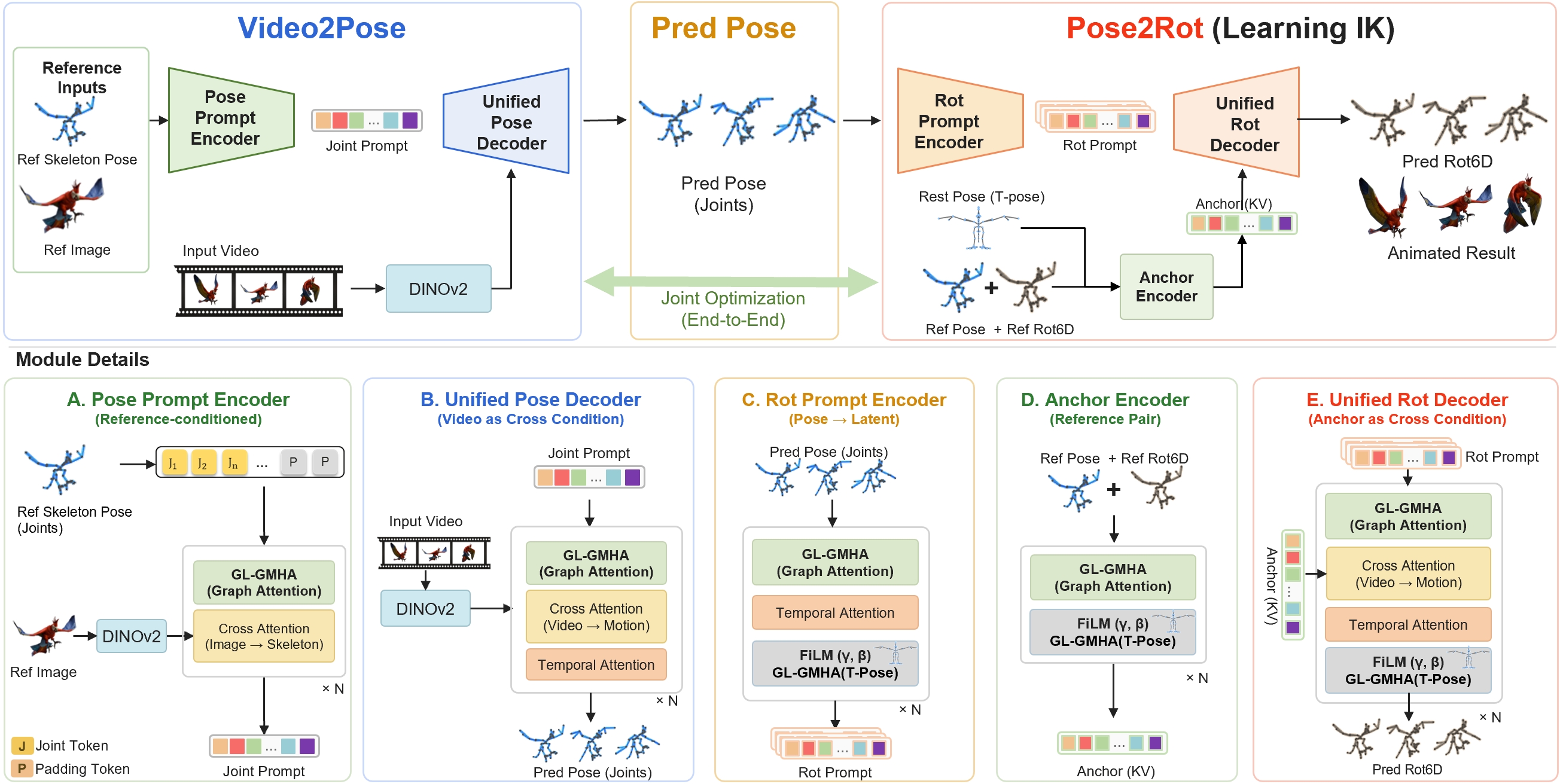}
  \caption{%
    Framework of MoCapAnything V2. Our method unifies video-to-pose and pose-to-rotation within a single end-to-end trainable architecture.
    The video-to-pose stage consists of a reference-conditioned pose prompt encoder (A), which encodes skeleton and image cues into joint prompt, and a unified pose decoder (B), which predicts temporally coherent joint positions via cross-attention with video features.
    The pose-to-rotation stage is formulated as a learnable inverse kinematics module, composed of a rotation prompt encoder (C) that maps predicted poses into rot prompt, an anchor encoder (D) that encodes reference pose--rotation pairs to establish a consistent rotation coordinate space, and a unified rotation decoder (E) that generates animation-ready joint rotations conditioned on the anchor through cross-attention.%
  }
  \Description{Detailed architecture diagram of MoCapAnything V2 with five labelled modules. The video-to-pose stage on the left contains a reference-conditioned pose prompt encoder (A) that fuses skeleton and image cues into a joint prompt, and a unified pose decoder (B) that produces per-frame joint positions via cross-attention with video features. The pose-to-rotation stage on the right contains a rotation prompt encoder (C) that maps predicted poses into a rotation prompt, an anchor encoder (D) that encodes a reference pose-rotation pair into a coordinate-system anchor, and a unified rotation decoder (E) that produces per-frame joint rotations conditioned on the anchor through cross-attention. Arrows indicate end-to-end gradient flow across both stages.}
  \label{fig:framework}
\end{figure*}

\subsection{Overview}
\label{sec:overview}

Unlike prior arbitrary-skeleton methods~\cite{mocapanythingv1}, which adopt a factorized design, a learned Video-to-Pose stage followed by a non-differentiable analytical IK stage for Pose-to-Rotation (See Fig.~\ref{fig:differencev1v2}), our framework casts \emph{both} stages as learnable neural modules and trains them jointly end-to-end (see Fig.~\ref{fig:framework}). This unified design is what makes joint optimization of the full V$\to$P$\to$R pipeline possible; its technical enabler is a reference pose--rotation pair that resolves the ill-posedness of a learnable P$\to$R mapping (\S\ref{sec:p2r}).

Concretely, our framework comprises two core modules:

\noindent\textbf{Video-to-Pose module} (\S\ref{sec:v2p}) predicts a sequence of joint positions from the input video, conditioned on a reference frame that establishes the skeleton-specific joint layout.

\noindent\textbf{Pose-to-Rotation module} (\S\ref{sec:p2r}) maps the predicted joint positions to per-joint rotations, conditioned on the target skeleton's rest pose and a reference pose--rotation pair that anchors the local coordinate system.

Both modules are conditioned on the \emph{same} reference frame, providing a consistent anchor throughout the pipeline: during training, this frame is randomly sampled from the ground-truth sequence, while at test time it is naturally provided together with the target skeleton asset (e.g., one frame from the rigged animation supplied with the asset).

\subsection{Global-Local Graph-Guided Multi-Head Attention}
\label{sec:gmha}

We adopt a topology-aware attention mechanism shared by both the Video-to-Pose and Pose-to-Rotation modules. Building on Graph-guided Multi-Head Attention (GMHA)~\cite{gat2025anytop}, which uses graph-derived joint relations (distance and kinematic connectivity) as attention bias, we introduce a \emph{Global-Local} variant (GL-GMHA) that alternates between local layers restricted to kinematic chains (capturing intra-limb dependencies) and global layers with full connectivity (capturing cross-limb coordination). This complementary design models both structural constraints and whole-body dynamics without additional parameters and naturally generalizes across diverse skeleton topologies.

\subsection{Video-to-Pose Module}
\label{sec:v2p}

This module predicts joint positions directly from the input video, without explicit geometric intermediates such as mesh or surface normals: predicted-mesh noise propagates through downstream stages and degrades stability (\S\ref{sec:exp_v1_compare}).

\paragraph{Reference query encoder.}
To ground the joint layout in the target skeleton, we encode a \emph{reference frame} for which joint positions and image features are both known. Reference joint positions $\mathbf{p}^{\text{ref}} \in \mathbb{R}^{J \times 3}$ are mapped through a frequency-based positional embedding~\cite{mildenhall2020nerf} and projected to dimension $d$. Per-joint semantic embeddings, obtained by encoding joint names with the T5~\cite{raffel2020exploring} text encoder, are added to provide category-agnostic joint identity that generalizes across arbitrary joint counts and naming conventions. Reference image features $\mathbf{z}^{\text{ref}} \in \mathbb{R}^{P \times d_{\text{img}}}$ (from a frozen DINOv2~\cite{oquab2023dinov2} encoder) are fused via a stack of \emph{RefFusionBlocks}, each comprising GL-GMHA self-attention over joints, vanilla self-attention, and cross-attention to image features. The output is a set of reference joint queries $\mathbf{Q}^{\text{ref}} \in \mathbb{R}^{J \times d}$ that encode both skeletal structure and reference appearance.

\paragraph{Temporal pose decoder.}
Given $\mathbf{Q}^{\text{ref}}$ and per-frame image features $\mathbf{z}_t$ (same frozen DINOv2 encoder), a temporal transformer uses GL-GMHA for spatial reasoning across joints and windowed per-joint temporal attention with RoPE~\cite{su2024roformer} across frames, yielding $\hat{\mathbf{P}} = \{\hat{\mathbf{p}}_t\} \in \mathbb{R}^{T \times J \times 3}$. Joints flagged as position-static during preprocessing are overwritten with the reference position to ensure structural consistency; an analogous rotation-static flag is handled in \S\ref{sec:rot_decoder}.

We treat joint positions as a \emph{skeleton-shared canonical representation}: different skeletons performing the same motion share similar position patterns, decoupling skeleton-specific conventions from motion content (\S\ref{sec:exp_pose_role}). Because P$\to$R is also learnable (\S\ref{sec:p2r}), this representation is co-adapted during end-to-end training (\S\ref{sec:training}) toward the final rotation objective rather than frozen by a positional-accuracy loss. Such co-adaptation is unattainable in factorized pipelines with non-differentiable IK.

\subsection{Reference-Conditioned Pose-to-Rotation Module}
\label{sec:p2r}

\subsubsection{The Ill-Posedness of Pose-to-Rotation Mapping}
\label{sec:illposed}

\begin{figure}[t]
    \centering
    \includegraphics[width=\columnwidth]{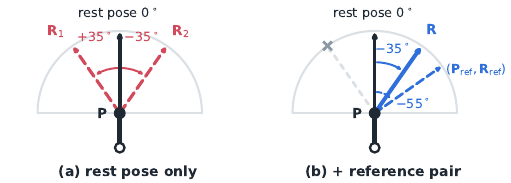}
        \caption{Why the rest pose alone does not determine joint rotations. (a) With the vertical rest-pose direction taken as $0^{\circ}$, the same positional constraint admits both $\mathbf{R}_1{=}{+}35^{\circ}$ and $\mathbf{R}_2{=}{-}35^{\circ}$ under different sign conventions. (b) A non-rest reference pair at $-55^{\circ}$ anchors the convention and selects $\mathbf{R}{=}{-}35^{\circ}$. The 2D diagram simplifies the corresponding 3D ambiguity, which also includes unconstrained bone-axis twist.}
    \Description{Two protractor-style diagrams. In the left panel a vertical arrow marks the rest-pose zero direction, and two dashed arrows at plus and minus thirty-five degrees show two rotations consistent with the same joint position. In the right panel a reference pose-rotation pair at minus fifty-five degrees selects the clockwise solution, and the counter-clockwise one is crossed out.}
    \label{fig:p2r_ambiguity}
\end{figure}

Joint rotations are defined relative to both a skeleton's rest pose and its local coordinate frames. The rest pose fixes the joint origins and bone directions, but does not fully determine the frame axes. As illustrated in Fig.~\ref{fig:p2r_ambiguity}(a), even in 2D the same positional constraint can admit rotations with opposite signs under different axis conventions. In 3D, positions additionally leave bone-axis twist entirely unconstrained. Consequently, conditioning only on the rest offsets $\mathbf{o}$ leaves $\mathbf{R}=f(\mathbf{P},\mathbf{o})$ multi-valued. A single non-rest reference pose--rotation pair resolves this ambiguity by demonstrating how one observed configuration is expressed in the target rig's convention (Fig.~\ref{fig:p2r_ambiguity}(b)), thereby anchoring the complete local coordinate system.

\subsubsection{Reference-Conditioned Modeling}
\label{sec:ref_cond}

We extend the static-geometry reference signal of prior work~\cite{mocapanythingv1} with a single pose--rotation pair sampled from the same asset. Intuitively, rest pose supplies the coordinate \emph{origin} while the reference pose--rotation pair supplies the coordinate \emph{axes}: together they fully specify the local frame convention, telling the model ``for this skeleton's coordinate definition, this joint configuration corresponds to these rotations,'' which converts the multi-valued P$\to$R mapping into a well-constrained conditional prediction task. We use a single reference pair by default; the effect of additional pairs is analyzed in our ablation study (\S\ref{sec:exp_ref_rest}). Concretely, the rotation prediction is conditioned on three encoders, all built from GL-GMHA layers:
(i) a \emph{Rest Pose Encoder} takes bone offsets $\mathbf{o} \in \mathbb{R}^{J \times 3}$ and per-joint semantic embeddings, producing a rest-pose feature $\mathbf{E}^{\text{rest}} \in \mathbb{R}^{J \times d}$ that captures static geometry and topology;
(ii) a \emph{Reference Encoder} jointly embeds the reference position $\mathbf{p}^{\text{ref}}$ and 6D rotation $\mathbf{r}^{\text{ref}}$, modulated by $\mathbf{E}^{\text{rest}}$ via FiLM~\cite{perez2018film}, producing the coordinate-system anchor $\mathbf{C}^{\text{ref}} \in \mathbb{R}^{J \times d}$;
(iii) a \emph{Pose Encoder} processes the predicted (or ground-truth) pose sequence $\mathbf{P}\in\mathbb{R}^{T\times J\times 3}$ via alternating GL-GMHA and per-joint windowed temporal attention with RoPE~\cite{su2024roformer}, optionally modulated by $\mathbf{E}^{\text{rest}}$, yielding a pose feature sequence $\mathbf{Q}\in\mathbb{R}^{T\times J\times d}$.

\subsubsection{Rotation Decoder}
\label{sec:rot_decoder}

The rotation decoder predicts per-frame, per-joint 6D rotations from the pose features $\mathbf{Q}$. Each of its $L$ blocks applies, in order: FiLM modulation by $\mathbf{E}^{\text{rest}}$ for skeleton-specific conditioning, per-joint temporal self-attention (windowed with RoPE) for temporal coherence, GL-GMHA spatial attention (alternating local/global masking) for cross-joint reasoning, per-joint cross-attention to the reference anchor $\mathbf{C}^{\text{ref}}$ (applied in the first $L_{\text{cross}}\leq L$ layers; the rest rely on already-integrated reference information), and a feed-forward residual; the final layer projects to 6D rotation via a two-layer MLP. Joints flagged as rotation-static (determined independently from the position-static flag in \S\ref{sec:v2p}) are overwritten with the reference rotation, mirroring the static-joint handling in the pose stage. By conditioning jointly on the rest pose, the reference pair, and the input pose, the decoder turns the ill-posed rotation recovery problem into a well-constrained conditional prediction task and learns statistical motion priors that resolve degrees of freedom (e.g., twist) inaccessible to analytical IK.

\subsection{End-to-End Training Strategy}
\label{sec:training}

\paragraph{Joint optimization.}
With P$\to$R cast as a learnable neural module (\S\ref{sec:p2r}), end-to-end joint training of the entire V$\to$P$\to$R pipeline becomes possible for the first time in arbitrary-skeleton motion capture. We train the Video-to-Pose and Pose-to-Rotation modules jointly: gradients from the rotation loss flow back through the predicted pose and into the visual encoder, so the intermediate pose representation is reshaped not by a positional-accuracy objective alone, but by what best serves the final rotation objective. This is the key departure from factorized pipelines, whose non-differentiable IK handoff between V$\to$P and P$\to$R precludes any such co-adaptation; we quantify the effect in \S\ref{sec:exp_E2Etraining}.

\paragraph{Loss function.}
The total loss combines four terms spanning both the pose and rotation stages:
\begin{equation}
  \mathcal{L} = \lambda_{\text{pos}} \, \mathcal{L}_{\text{pos}}
              + \lambda_{\text{rot}} \, \mathcal{L}_{\text{rot}}
              + \lambda_{\text{rot\_v}} \, \mathcal{L}_{\text{rot\_v}}
              + \lambda_{\text{root}} \, \mathcal{L}_{\text{root}},
  \label{eq:loss}
\end{equation}
where $\mathcal{L}_{\text{pos}}$ is the per-joint position error between predicted and ground-truth joint positions. For the rotation stage, $\mathcal{L}_{\text{rot}}$ measures the geodesic angular error between predicted and ground-truth rotations averaged over all joints, and $\mathcal{L}_{\text{rot\_v}}$ penalizes the angular velocity difference to promote temporal consistency in the rotation sequence. $\mathcal{L}_{\text{root}}$ additionally re-weights the root joint's rotation error, which we found accelerates convergence of the global orientation. All losses are computed with per-joint masking to handle the variable number of joints across different skeletons. We set $\lambda_{\text{pos}}{=}\lambda_{\text{rot}}{=}\lambda_{\text{rot\_v}}{=}1.0$ and $\lambda_{\text{root}}{=}0.1$.

\paragraph{Mixed-pose training.}
A distribution gap exists between training, where the Pose-to-Rotation module can receive ground-truth poses, and inference, where it must operate on noisy predicted poses. To bridge this gap, we employ a mixed-pose training strategy that stochastically selects, for each sample in a batch, whether to feed ground-truth or predicted poses to the rotation module. The probability of using predicted poses follows a schedule:
\begin{equation}
  p_{\text{pred}}(e) = p_{\text{start}} + (p_{\text{end}} - p_{\text{start}}) \cdot \min\!\Big(1,\; \frac{e}{E_{\text{warmup}}}\Big),
  \label{eq:mix_schedule}
\end{equation}
where $e$ is the current epoch and $E_{\text{warmup}}$ controls the transition rate. We set $p_{\text{start}}{=}0.1$ and $p_{\text{end}}{=}1.0$, so that early in training ground-truth poses dominate to ensure stable convergence, and the proportion of predicted poses increases gradually until the model is fully trained on its own pose predictions by the end of the warm-up phase. The choice of $E_{\text{warmup}}$ is studied in Appendix~\ref{sec:appendix_warmup}.

\section{Experiments}
\label{sec:experiments}

\subsection{Dataset and Evaluation Protocol}
\label{sec:exp_setup}

\paragraph{Datasets.}
We evaluate on two benchmarks spanning diverse skeleton structures:
\begin{itemize}
    \item \textbf{Truebones Zoo}~\cite{truebones_mocap}: 1{,}038 animal motion sequences (104{,}715 frames) covering a broad range of species and kinematic topologies. The test set (60 sequences) is stratified into \textbf{Seen} (species with abundant training data), \textbf{Rare} (species with limited training data), and \textbf{Unseen} (species never seen during training).
    \item \textbf{Objaverse}: 1{,}000 samples from Objaverse~\cite{objaverse,objaverseXL}, containing structurally distinct humanoid and non-animal targets unseen during training, serving as an out-of-distribution stress test.
\end{itemize}
We additionally evaluate on \textbf{in-the-wild videos} collected from the Internet to assess real-world robustness.

\paragraph{Evaluation metrics.}
We report four metrics spanning spatial accuracy and rotation quality:
\textbf{MPJPE} (Mean Per Joint Position Error, cm),
\textbf{MPJVE} (Mean Per Joint Velocity Error, cm),
\textbf{Ang.\ Err} (geodesic angle error, $^{\circ}$), and
\textbf{AngV Err} (angular velocity error, $^{\circ}$).
To handle large inter-species scale variations, all samples are normalized to $[-1,1]^3$ for training and rescaled to a unified $1\,\text{m}^3$ cube for evaluation. All metrics are computed with per-joint masking to accommodate varying joint counts.

\paragraph{Baselines.}
We compare against \textbf{HRNet}~\cite{sun2019deep}, \textbf{ViTPose}~\cite{xu2022vitpose}, \textbf{VIBE}~\cite{kocabas2020vibe}, and \textbf{GLoT}~\cite{shen2023global}. For each baseline, we instantiate both the Video-to-Pose and Pose-to-Rotation modules with that method's architecture and train them jointly end-to-end on the same training set with unified skeleton representations. Our most direct point of comparison is \textbf{MoCapAnything V1}~\cite{mocapanythingv1}, which shares the same problem setting and training data, and differs only in that V1 adopts a factorized design: a learned Video-to-Pose stage with a 4D mesh intermediate, followed by a constraint-aware analytical IK stage for Pose-to-Rotation, with the two stages optimized independently.
All learned baselines use the same frozen DINO features, data, training epochs, aligned inputs, unified skeleton representation, and end-to-end V$\to$P$\to$R interface. GLoT originally assumes a fixed 24-joint SMPL topology and dense meshes; because neither is shared across arbitrary skeletons, we replace these human-specific components with the topology-general interface used by the other baselines.

\paragraph{Implementation details.}
We use a frozen DINOv2~\cite{oquab2023dinov2} ViT encoder as the visual backbone and jointly train the Video-to-Pose and Pose-to-Rotation modules end-to-end. The rotation decoder has $L{=}8$ blocks, with reference cross-attention in the first $L_{\text{cross}}{=}6$ layers. We process $T{=}48$-frame sequences with a per-joint temporal window of $5$. Training uses Adam on $8{\times}$ V100 GPUs for $60$ epochs with batch size $8$ and takes roughly one day. We set the loss weights $\lambda_{\text{pos}}{=}\lambda_{\text{rot}}{=}\lambda_{\text{rot\_v}}{=}1.0$ and $\lambda_{\text{root}}{=}0.1$, and the mixed-pose schedule parameters $(p_{\text{start}},p_{\text{end}},E_{\text{warmup}}){=}(0.1,1.0,30)$. The maximum joint count is $150$ (the largest training skeleton has $143$ joints) and can be adjusted for other datasets.

\subsection{Comparison with Baselines}
\label{sec:exp_main}

Table~\ref{tab:exp_main} presents a comprehensive comparison across all four evaluation splits.


\begin{table}[t]
\centering
\caption{Main results on Zoo (Seen/Rare/Unseen) and Obj. Position in cm ($\downarrow$); rotation in degrees ($\downarrow$). All baselines are trained jointly end-to-end with learnable rotation modules; only V1 uses traditional IK. Best angle error per split in \textbf{bold}.}
\label{tab:exp_main}
\resizebox{\columnwidth}{!}{
\begin{tabular}{l|cccc|cccc|cccc|cccc}
\toprule
& \multicolumn{4}{c|}{\textbf{Zoo-Seen}}
& \multicolumn{4}{c|}{\textbf{Zoo-Rare}}
& \multicolumn{4}{c|}{\textbf{Zoo-Unseen}}
& \multicolumn{4}{c}{\textbf{Obj}} \\
\cmidrule(lr){2-5}\cmidrule(lr){6-9}\cmidrule(lr){10-13}\cmidrule(lr){14-17}
Method
 & JP & JV & An & AV
 & JP & JV & An & AV
 & JP & JV & An & AV
 & JP & JV & An & AV \\
\midrule
HRNet ($lr{=}5{\times}10^{-5}$) & 19.83 & 0.56 & 19.67 & 0.51 & 20.51 & 0.68 & 24.38 & 0.63 & 25.77 & 1.45 & 24.56 & 0.64 & 23.43 & 1.55 & 31.11 & 0.75 \\
GLoT & 19.66 & 0.60 & 20.24 & 0.52 & 21.60 & 0.70 & 26.13 & 0.65 & 26.13 & 1.47 & 25.95 & 0.66 & 22.95 & 1.70 & 29.07 & 0.69 \\
ViTPose & 19.77 & 0.59 & 20.90 & 0.52 & 21.12 & 0.68 & 25.48 & 0.63 & 26.17 & 1.45 & 24.46 & 0.63 & 24.16 & 1.85 & 29.30 & 0.70 \\
VIBE & 19.69 & 0.54 & 19.67 & 0.51 & 20.77 & 0.63 & 25.06 & 0.63 & 26.29 & 1.45 & 25.74 & 0.65 & 23.51 & 1.63 & 28.72 & 0.70 \\
\midrule
\textbf{Ours}
& 2.34 & 0.53 & \textbf{10.73} & 0.29 & 2.98 & 0.61 & \textbf{14.38} & 0.37 & 3.39 & 0.99 & \textbf{6.54} & 0.17 & 3.84 & 1.05 & \textbf{11.06} & 0.30 \\
\bottomrule
\end{tabular}
}
\end{table}


Our method achieves consistently strong results across all splits, with the largest gains on rotation. Although the learned baselines receive the same reference inputs and adopt an end-to-end V$\to$P$\to$R formulation, their architectures do not effectively leverage reference and topology cues to resolve coordinate-axis ambiguity on arbitrary skeletons, capping rotation quality near $20^{\circ}$ angle error with artifacts such as joint spinning. Our reference-conditioned design halves this to ${\sim}10^{\circ}$ with substantially lower angular velocity error. The improvement is most pronounced on Zoo-Unseen, where the angle error ($6.54^{\circ}$) is actually \emph{lower} than Zoo-Seen ($10.73^{\circ}$) and Zoo-Rare ($14.38^{\circ}$): this split is dominated by common locomotion motions (\eg, walking, running) whose rotations become straightforward once the coordinate axes are anchored by the reference pair. A shared learning rate of $10^{-4}$ was suitable for the Transformer baselines but made HRNet's position head unstable. Sweeping $\{10^{-6},10^{-5},3{\times}10^{-5},5{\times}10^{-5}\}$ identified $5{\times}10^{-5}$ as stable, reducing HRNet's Zoo-Unseen MPJPE from $37.87$ to $25.77$ cm, while ViTPose remains at $26.17$ cm. HRNet therefore matches ViTPose, confirming that the gap reflects learning-rate sensitivity rather than a structural issue with the comparison. V1 is compared separately in \S\ref{sec:exp_v1_compare}. A motion-category breakdown for unseen skeletons, reference sensitivity analysis, analytical IK alternatives, and joint-name embedding ablation are reported in \S\ref{sec:additional_diagnostics}.

\begin{figure*}[!t]
    \centering
    \includegraphics[width=\textwidth]{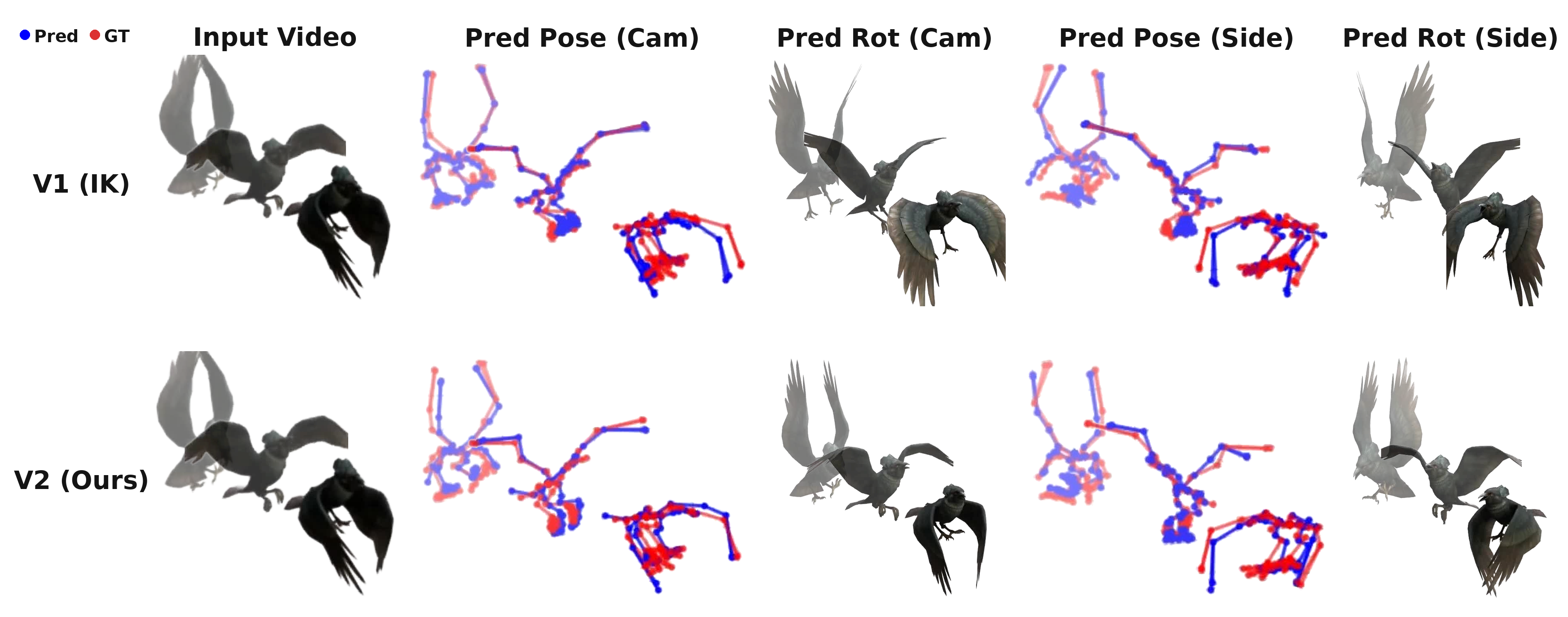}
    \caption{\textbf{MoCap V1 vs.\ V2.} Row~1: V1 (traditional IK-based optimization). Row~2: V2 (our learning-based rotation recovery). Each cell stacks three keyframes (diagonal offset, alpha fade) to convey motion. V1 suffers from joint spinning artifacts visible across the ghosted frames, whereas V2 produces stable, temporally consistent rotations.}
    \Description{Static motion-ghost qualitative comparison on a bird skeleton across two rows. Row 1 shows V1 results from traditional analytical IK, exhibiting visible joint spinning and limb flipping artifacts during flight. Row 2 shows V2 results from our learning-based rotation decoder, with smooth and temporally consistent wing and limb rotations. Each cell stacks three keyframes of the wing-flap cycle.}
    \label{fig:mocap_compare}
\end{figure*}
\subsection{End-to-End vs.\ Factorized Design: Comparison with V1}
\label{sec:exp_v1_compare}

V1~\cite{mocapanythingv1} differs from our method along two simultaneous axes: (i) presence of a 4D mesh intermediate between video and pose, and (ii) a factorized learned-V$\to$P + analytical-IK P$\to$R design versus our end-to-end learnable V$\to$P$\to$R. We compare against V1 under different mesh configurations in Table~\ref{tab:exp_v1_compare} to disentangle the two effects.


\begin{table}[t]
\centering
\caption{V1 vs.\ Ours under different mesh configurations. ``GT Mesh'' = ground-truth mesh; ``Pred Mesh'' = predicted mesh; ``Ours'' removes mesh entirely. For fair comparison with V1 (which was trained on Zoo only), all models in this table are both trained and evaluated on Zoo only; the Ours numbers therefore differ slightly from those in Table~\ref{tab:exp_main}, where the model is trained on Zoo+Obj. Best angle error per split in \textbf{bold}.}
\label{tab:exp_v1_compare}
\resizebox{\columnwidth}{!}{
\begin{tabular}{l|cccc|cccc|cccc}
\toprule
& \multicolumn{4}{c|}{\textbf{Zoo-Seen}}
& \multicolumn{4}{c|}{\textbf{Zoo-Rare}}
& \multicolumn{4}{c}{\textbf{Zoo-Unseen}} \\
\cmidrule(lr){2-5}\cmidrule(lr){6-9}\cmidrule(lr){10-13}
Config
 & JP & JV & An & AV
 & JP & JV & An & AV
 & JP & JV & An & AV \\
\midrule
V1 (GT Mesh+IK)
 & 1.06 & 0.44 & 17.47 & 2.18 & 1.28 & 0.37 & 18.52 & 2.08 & 1.76 & 0.36 & 20.56 & 3.10 \\
V1 (Pred Mesh+IK)
 & 3.30 & 0.67 & 20.02 & 2.51 &  4.19 & 0.55 & 19.82 & 2.18 & 4.78 & 0.93 & 22.04 & 3.37 \\
\midrule
\textbf{Ours}
 & 2.20 & 0.53 & \textbf{10.91} & 0.30
 & 2.86 & 0.59 & \textbf{14.36} & 0.37
 & 3.73 & 0.98 & \textbf{6.68} & 0.18 \\
\bottomrule
\end{tabular}
}
\end{table}
V1 with ground-truth mesh achieves the lowest positional error, confirming geometry helps \emph{when accurate}; but GT mesh is unavailable at inference. With predicted mesh (official implementation), V1's factorized design propagates noise through the mesh bridge, performing markedly worse than our mesh-free model. Our end-to-end V$\to$P$\to$R reaches pose accuracy competitive with V1 (GT Mesh) without any mesh, and substantially outperforms all V1 variants on rotation, reducing angle error from V1's $17^{\circ}$--$22^{\circ}$ range (with frequent joint spinning and limb flipping on twist-heavy joints; see Fig.~\ref{fig:mocap_compare}) to ${\sim}10^{\circ}$ with much lower angular velocity. Two factors drive this gain, analyzed next: end-to-end joint training adapts the pose to the rotation objective (\S\ref{sec:exp_E2Etraining}), and the reference pose--rotation pair anchors the coordinate system to make learnable P$\to$R feasible (\S\ref{sec:exp_ref_rest}). A stronger mesh predictor could narrow the positional gap but at significant overhead (\S\ref{sec:exp_efficiency}).

\subsection{Training Strategy Analysis: Validating the End-to-End Claim}
\label{sec:exp_E2Etraining}

This experiment directly tests the central claim of our framework: making P$\to$R learnable is not sufficient on its own; its benefit arises from enabling end-to-end gradient coupling with V$\to$P. We therefore compare four training regimes: a gradient-detached variant (where gradients from P$\to$R do not flow back to V$\to$P), end-to-end training with ground-truth poses only, end-to-end training with predicted poses only, and our default mixed-pose end-to-end training. Results are shown in Table~\ref{tab:exp_E2Etraining}.

\begin{table}[t]
\centering
\caption{Ablation of training strategies on Zoo (Seen/Rare/Unseen) and Obj. Position in cm; rotation in degrees ($\downarrow$). Best angle error per split in \textbf{bold}.}
\label{tab:exp_E2Etraining}
\resizebox{\columnwidth}{!}{
\begin{tabular}{l|cccc|cccc|cccc|cccc}
\toprule
& \multicolumn{4}{c|}{Zoo-Seen}
& \multicolumn{4}{c|}{Zoo-Rare}
& \multicolumn{4}{c|}{Zoo-Unseen}
& \multicolumn{4}{c}{Obj} \\
\cmidrule(lr){2-5}\cmidrule(lr){6-9}\cmidrule(lr){10-13}\cmidrule(lr){14-17}
Strategy
 & JP & JV & An & AV
 & JP & JV & An & AV
 & JP & JV & An & AV
 & JP & JV & An & AV \\
\midrule
Mixed (gradient detached) (Zoo only) & 1.76 & 0.39 & 11.69 & 0.32 & 2.50 & 0.50 & 15.75 & 0.42 & 2.90 & 0.79 & 6.60 & 0.18 & -- & -- & -- & -- \\  
Mixed (gradient detached)
 & 2.05 & 0.43 & 11.67 & 0.32
 & 2.43 & 0.50 & 14.80 & 0.38
 & 2.91 & 0.89 & 7.82 & 0.21
 & 3.96 & 1.01 & 11.96 & 0.29 \\
GT pose only
 & 2.10 & 0.43 & 12.68 & 0.35
 & 2.71 & 0.55 & 14.74 & 0.39
 & 3.39 & 0.86 & 13.28 & 0.38
 & 3.59 & 1.01 & 16.39 & 0.41 \\
Pred pose only
 & 3.47 & 0.65 & 11.91 & 0.32
 & 4.23 & 0.70 & 15.32 & 0.39
 & 4.42 & 1.14 & 9.58 & 0.25
 & 4.57 & 1.18 & 11.75 & 0.30 \\
Mixed (with joint opt. ours) & 2.34 & 0.53 & \textbf{10.73} & 0.29 & 2.98 & 0.61 & \textbf{14.38} & 0.37 & 3.39 & 0.99 & \textbf{6.54} & 0.17 & 3.84 & 1.05 & \textbf{11.06} & 0.30 \\
\bottomrule
\end{tabular}
}
\end{table}

Enabling gradient flow is critical: joint training reduces the Zoo-Unseen angle error from $7.82^{\circ}$ (detached) to $6.54^{\circ}$, showing the benefit of a learnable P$\to$R module comes from end-to-end adaptation rather than standalone modeling. GT-only training suffers a distribution gap and fails on Zoo-Unseen ($13.28^{\circ}$), while Pred-only is unstable due to noisy early predictions; the mixed-pose schedule achieves the best balance (warm-up sensitivity in Appendix~\ref{sec:appendix_warmup}).

\subsection{Pose-to-Rotation Module Analysis}
\label{sec:exp_ref_rest}

\paragraph{Reference pair and rest pose.}
Table~\ref{tab:exp_ref_rest} examines the contributions of the reference pose--rotation pair and the rest-pose encoding.

\begin{table}[t]
\centering
\caption{Ablation of reference conditioning (Ref) and rest pose (Rest). Position in cm; rotation in degrees ($\downarrow$). Best angle error per split in \textbf{bold}.}
\label{tab:exp_ref_rest}
\resizebox{\columnwidth}{!}{
\begin{tabular}{cc|cccc|cccc|cccc|cccc}
\toprule
& & \multicolumn{4}{c|}{\textbf{Zoo-Seen}}
    & \multicolumn{4}{c|}{\textbf{Zoo-Rare}}
    & \multicolumn{4}{c|}{\textbf{Zoo-Unseen}}
    & \multicolumn{4}{c}{\textbf{Obj}} \\
\cmidrule(lr){3-6}\cmidrule(lr){7-10}\cmidrule(lr){11-14}\cmidrule(lr){15-18}
Ref & Rest
 & JP & JV & An & AV
 & JP & JV & An & AV
 & JP & JV & An & AV
 & JP & JV & An & AV \\
\midrule
-- & -- & 2.23 & 0.54 & 11.25 & 0.31 & 2.97 & 0.61 & 14.83 & 0.40 & 3.07 & 0.98 & 24.26 & 0.64 & 4.22 & 1.10 & 18.00 & 0.46 \\
-- & \cmark & 2.24 & 0.53 & 11.15 & 0.31 & 3.11 & 0.61 & \textbf{13.92} & 0.37 & 3.18 & 0.96 & 24.05 & 0.64 & 3.61 & 1.05 & 16.14 & 0.41 \\
\cmark & -- & 2.39 & 0.56 & 11.31 & 0.31 & 2.98 & 0.59 & 14.95 & 0.39 & 3.22 & 0.98 & 7.37 & 0.19 & 3.71 & 1.08 & \textbf{11.02} & 0.27 \\
\cmark & \cmark
& 2.34 & 0.53 & \textbf{10.73} & 0.29 & 2.98 & 0.61 & 14.38 & 0.37 & 3.39 & 0.99 & \textbf{6.54} & 0.17 & 3.84 & 1.05 & 11.06 & 0.30 \\
\bottomrule
\end{tabular}
}
\end{table}

This empirically confirms the \emph{origin/axes} decomposition of \S\ref{sec:illposed}. On Zoo-Seen and Zoo-Rare, whose axis conventions are well represented in training, all four configurations reach comparable angle errors in the $10$--$15^{\circ}$ range: the axis convention is effectively memorized, so the explicit anchor offers little extra gain. The picture changes dramatically on Zoo-Unseen: without the reference pair, error jumps to $24.05^{\circ}$--$24.26^{\circ}$ (rest pose supplies only the coordinate origin, leaving axes ambiguous); adding the reference pair drops it to $7.37^{\circ}$, and combining both yields the best $6.54^{\circ}$. Rest pose thus provides a smaller but consistent benefit on top of the reference as complementary structural context.

\begin{figure*}[!t]
    \centering
    \includegraphics[width=\textwidth]{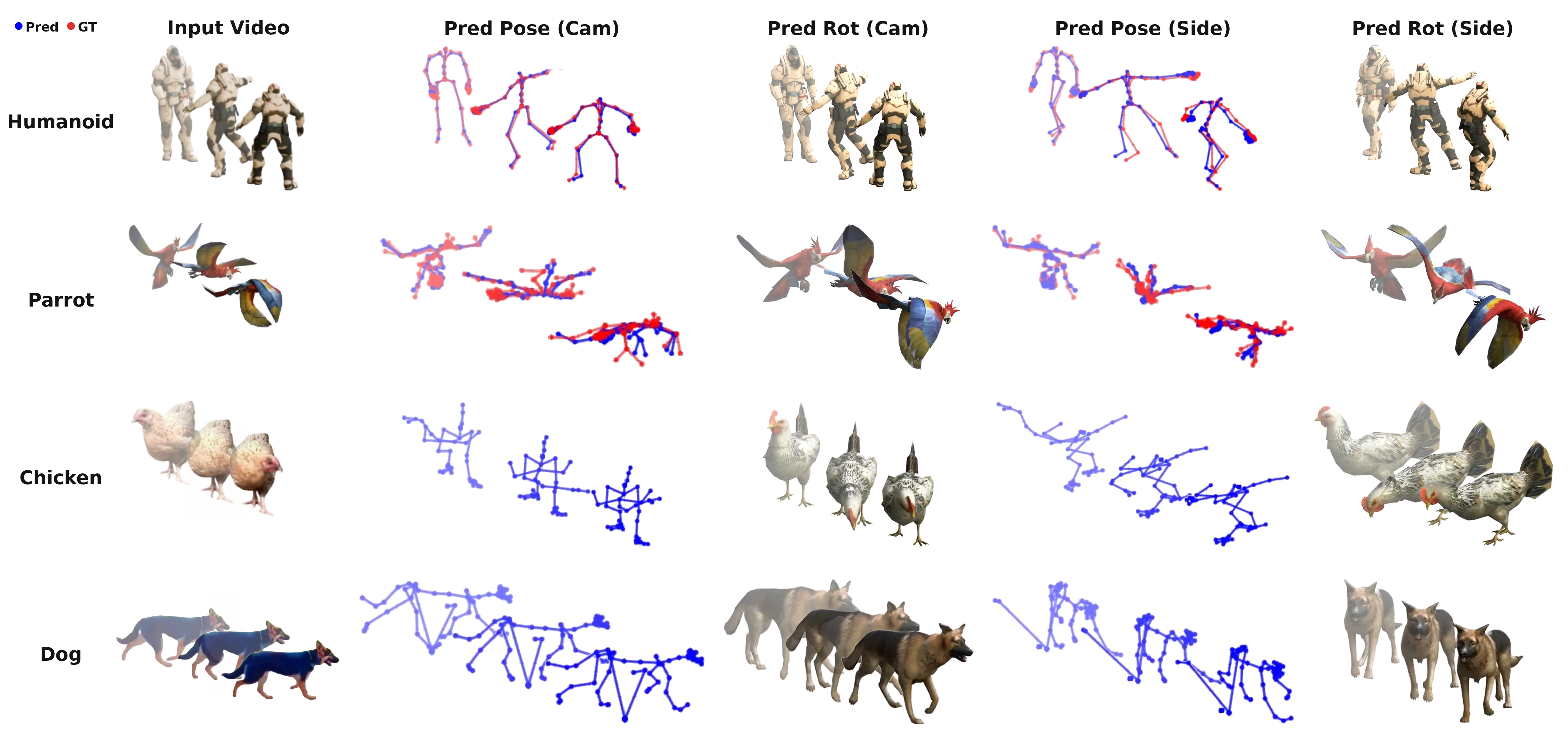}
    \caption{\textbf{MoCap demo across domains.} Row~1: Objaverse assets; Row~2: Truebones Zoo; Rows~3--4: in-the-wild videos. Each cell stacks three keyframes (diagonal offset, alpha fade) to convey motion across the sequence. Results are shown from multiple viewpoints, demonstrating accurate mocap on arbitrary subjects.}
    \Description{Static motion-ghost qualitative grid of motion capture results across three domains rendered from multiple viewpoints. Row 1 shows reconstructed motion for Objaverse humanoid and non-animal assets. Row 2 shows reconstructed motion for Truebones Zoo animal skeletons. Rows 3 and 4 show reconstructed motion driven by in-the-wild Internet videos. Each cell stacks three keyframes spaced across the motion.}
    \label{fig:mocap_demo}
\end{figure*}

\begin{figure*}[!t]
    \centering
    \includegraphics[width=\textwidth]{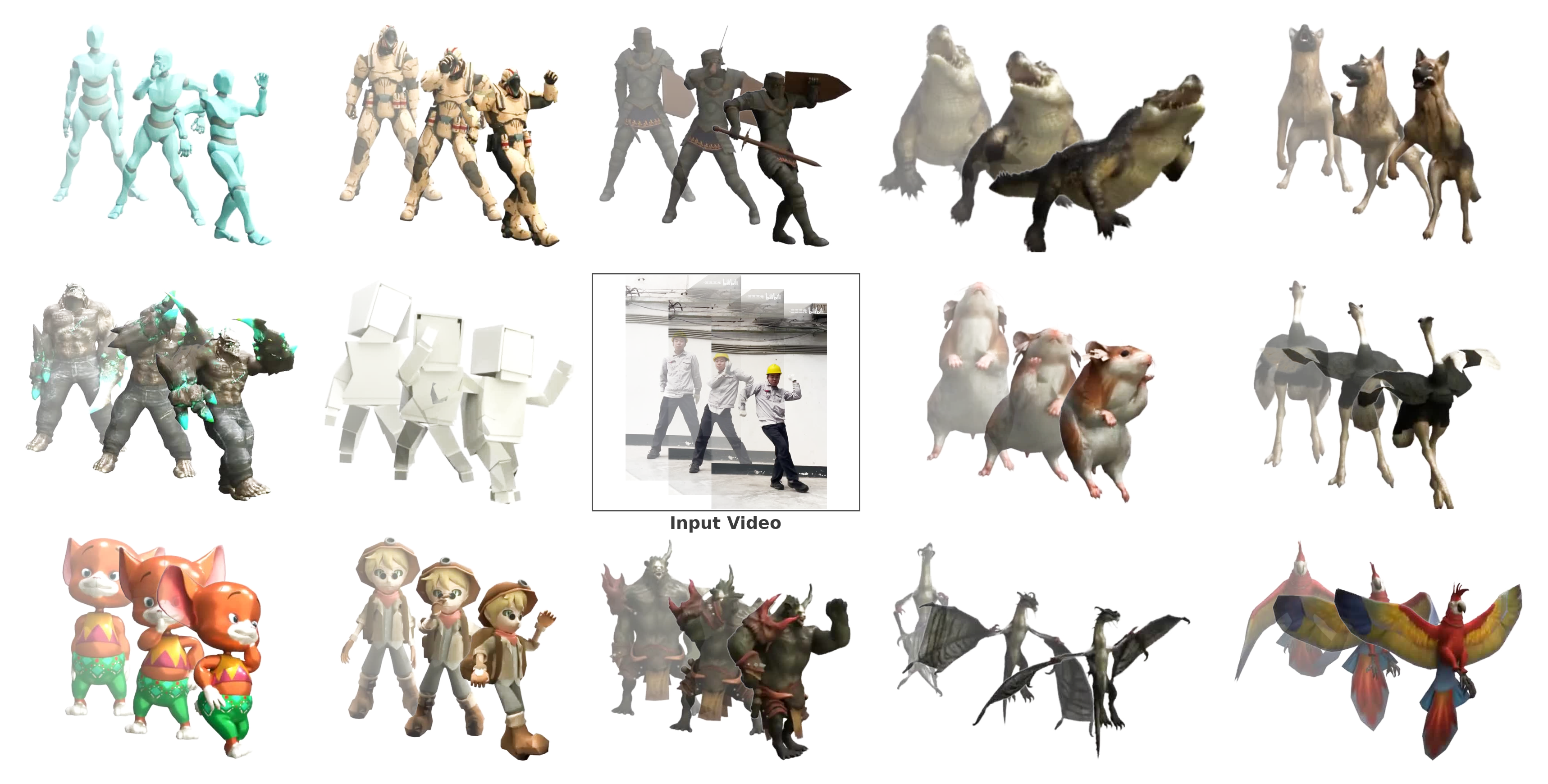}
    \caption{\textbf{Dance demo A.} A single input video (center) is mocapped and retargeted to a diverse set of humanoid and animal skeletons, all driven from the same source motion. Each cell stacks three keyframes (diagonal offset, alpha fade) to convey motion.}
    \Description{Static motion-ghost grid of 3 rows and 5 columns. The center cell is the input dance video. The other 14 cells are the same motion retargeted onto a variety of humanoid and animal characters, each shown as a 3-keyframe motion ghost.}
    \label{fig:dance_demoA}
\end{figure*}

\begin{figure*}[!t]
    \centering
    \includegraphics[width=\textwidth]{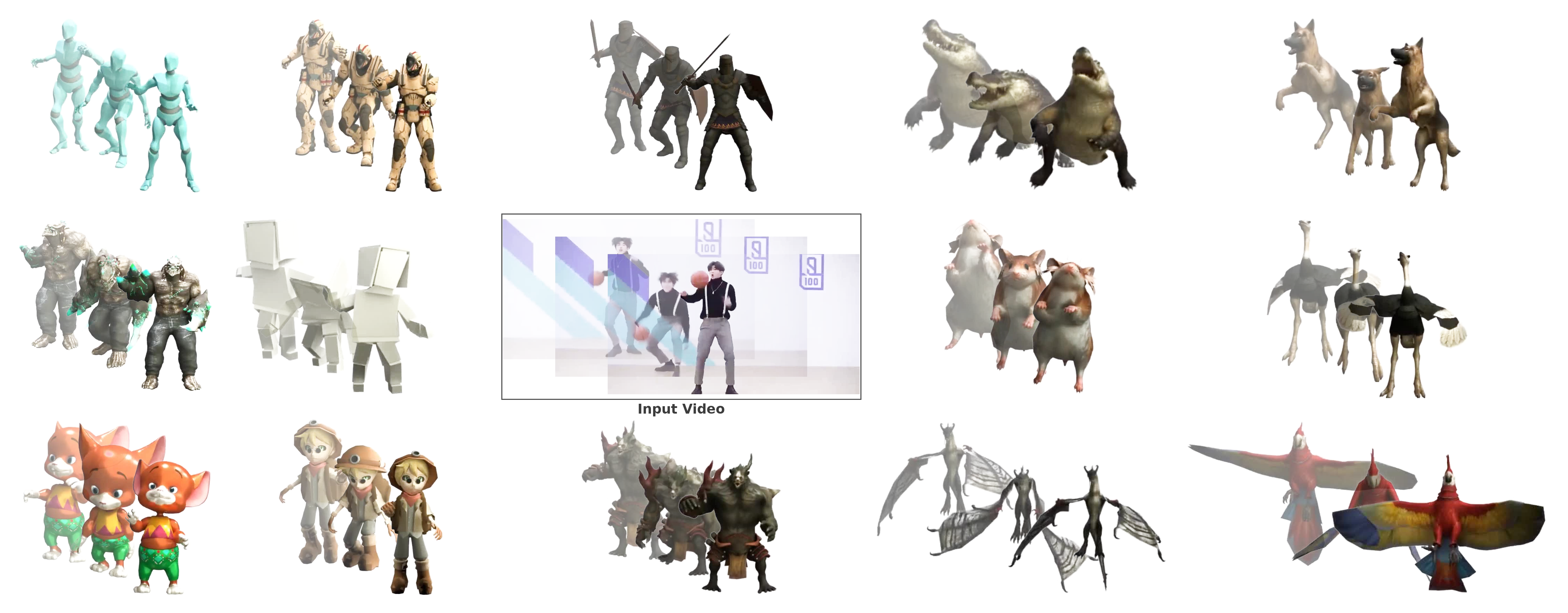}
    \caption{\textbf{Dance demo B.} A single input video (center) is mocapped and retargeted to a diverse set of humanoid and animal skeletons, all driven from the same source motion. Each cell stacks three keyframes (diagonal offset, alpha fade) to convey motion.}
    \Description{Static motion-ghost grid of 3 rows and 5 columns. The center cell is the input dance video (a second source clip). The other 14 cells are the same motion retargeted onto a variety of humanoid and animal characters, each shown as a 3-keyframe motion ghost.}
    \label{fig:dance_demoB}
\end{figure*}

\subsection{Role of Pose as Intermediate Representation}
\label{sec:exp_pose_role}

To validate the necessity of explicit joint positions as an intermediate representation, we compare three architectural variants in Table~\ref{tab:exp_pose_role}.


\begin{table}[t]
\centering
\caption{Ablation of the intermediate pose representation. Position in cm; rotation in degrees ($\downarrow$). ``Direct (V$\to$R)'' regresses rotations directly from video without a pose branch; joint positions for this variant are recovered by applying forward kinematics to the predicted rotations. ``Latent + Aux'' uses a latent intermediate supervised by an auxiliary pose loss. ``Full'' (Ours) uses an explicit joint-position intermediate as the first-stage output. Best angle error per split in \textbf{bold}.}
\label{tab:exp_pose_role}
\resizebox{\columnwidth}{!}{
\begin{tabular}{l|cccc|cccc|cccc|cccc}
\toprule
& \multicolumn{4}{c|}{\textbf{Zoo-Seen}}
& \multicolumn{4}{c|}{\textbf{Zoo-Rare}}
& \multicolumn{4}{c|}{\textbf{Zoo-Unseen}}
& \multicolumn{4}{c}{\textbf{Obj}} \\
\cmidrule(lr){2-5}\cmidrule(lr){6-9}\cmidrule(lr){10-13}\cmidrule(lr){14-17}
Architecture
 & JP & JV & An & AV
 & JP & JV & An & AV
 & JP & JV & An & AV
 & JP & JV & An & AV \\
\midrule
Direct (V$\to$R) & -- & -- & 9.32 & 0.26 & -- & -- & 12.71 & 0.34 & -- & -- & 23.73 & 0.62 & -- & -- & 11.16 & 0.28 \\
Latent + Aux & 2.07 & 0.47 & \textbf{9.06} & 0.25 & 2.40 & 0.53 & \textbf{11.85} & 0.31 & 2.88 & 0.89 & 23.57 & 0.62 & 3.82 & 1.04 & \textbf{10.71} & 0.27 \\
\textbf{Full (explicit pose, Ours)}
& 2.34 & 0.53 & 10.73 & 0.29 & 2.98 & 0.61 & 14.38 & 0.37 & 3.39 & 0.99 & \textbf{6.54} & 0.17 & 3.84 & 1.05 & 11.06 & 0.30 \\
\bottomrule
\end{tabular}
}
\end{table}

Both Direct (V$\to$R, $23.73^{\circ}$) and Latent+Aux ($23.57^{\circ}$) match our Seen/Rare performance ($9.32^{\circ}$/$12.71^{\circ}$ and $9.06^{\circ}$/$11.85^{\circ}$ respectively) but collapse on Zoo-Unseen, indicating that without an explicit pose intermediate, even when the latent pose is supervised, the model fails to generalize across unseen skeleton topologies. Our explicit-pose intermediate substantially improves Zoo-Unseen ($6.54^{\circ}$) while remaining competitive on Seen/Rare. This confirms that joint positions are skeleton-shared while rotations are skeleton-dependent: the explicit pose acts as a structural bottleneck that separates transferable motion patterns from skeleton-specific parameterization.

\subsection{GL-GMHA Attention Analysis}
\label{sec:exp_gmha}

Table~\ref{tab:exp_gmha} evaluates the effectiveness of the Global-Local extension to graph-guided attention.

\begin{table}[t]
\centering
\caption{Ablation of attention mechanisms. Position in cm; rotation in degrees ($\downarrow$). All variants use the same backbone; only the spatial attention pattern differs. ``GMHA (all-global)'' is the original uniform formulation from~\cite{gat2025anytop} in which every layer attends over all joints; ``All-local'' applies the ancestor mask at every layer; ``GL-GMHA (Ours)'' alternates between local and global layers. Best angle error per split in \textbf{bold}.}
\label{tab:exp_gmha}
\resizebox{\columnwidth}{!}{
\begin{tabular}{l|cccc|cccc|cccc|cccc}
\toprule
& \multicolumn{4}{c|}{\textbf{Zoo-Seen}}
& \multicolumn{4}{c|}{\textbf{Zoo-Rare}}
& \multicolumn{4}{c|}{\textbf{Zoo-Unseen}}
& \multicolumn{4}{c}{\textbf{Obj}} \\
\cmidrule(lr){2-5}\cmidrule(lr){6-9}\cmidrule(lr){10-13}\cmidrule(lr){14-17}
Attention
 & JP & JV & An & AV
 & JP & JV & An & AV
 & JP & JV & An & AV
 & JP & JV & An & AV \\
\midrule
Full Attn (no graph bias) & 2.83 & 0.48 & 12.57 & 0.33 & 3.57 & 0.57 & 16.11 & 0.41 & 3.98 & 1.01 & 11.92 & 0.31 & 4.64 & 1.07 & 12.02 & 0.30 \\  
GMHA (all-global)~\cite{gat2025anytop} & 2.17 & 0.53 & 11.18 & 0.31 & 3.17 & 0.65 & 14.51 & 0.39 & 3.46 & 0.98 & 6.69 & 0.18 & 3.76 & 1.08 & 11.21 & 0.28 \\
All-local (ancestor mask every layer) & 2.47 & 0.55 & 12.55 & 0.33 & 3.09 & 0.65 & 16.97 & 0.43 & 3.39 & 1.02 & 11.60 & 0.31 & 4.28 & 1.10 & 13.59 & 0.34 \\
\textbf{GL-GMHA (Ours)} & 2.34 & 0.53 & \textbf{10.73} & 0.29 & 2.98 & 0.61 & \textbf{14.38} & 0.37 & 3.39 & 0.99 & \textbf{6.54} & 0.17 & 3.84 & 1.05 & \textbf{11.06} & 0.30 \\
\bottomrule
\end{tabular}
}
\end{table}

The ``All-local'' variant, which restricts attention to ancestor paths at every layer, performs worst overall: it captures intra-branch dependencies but loses cross-branch coordination, with angle error jumping to $16.97^{\circ}$ on Zoo-Rare and $11.60^{\circ}$ on Zoo-Unseen. Removing graph biases entirely (``Full Attn'') trails uniform GMHA~\cite{gat2025anytop}, indicating that the structural inductive bias on joint relationships matters; uniform GMHA attends globally in every layer but does not explicitly model kinematic-chain locality. Our alternating GL-GMHA consistently outperforms uniform GMHA across all four splits, validating that kinematic-chain locality and skeleton-global coordination are \emph{complementary} rather than redundant.

\subsection{Additional Diagnostic Analyses}
\label{sec:additional_diagnostics}

\paragraph{Joint-name embeddings.}
Removing the T5 joint-name embedding changes angular error by about $0.5^{\circ}$ or less across all splits, which is within run-to-run variation. Joint names are generated automatically from the rendered skeleton and joint markers; missing or inconsistent names therefore degrade gracefully rather than becoming a manual requirement.

\paragraph{Diagnostic protocol.}
The motion-difficulty, reference-sensitivity, and P$\to$R-isolation analyses below use an independently retrained diversified held-out setting in which the evaluated six species are excluded from training. These absolute values are therefore not directly comparable with Table~\ref{tab:exp_main}; they are used only for the three diagnostic questions examined here.

\paragraph{Motion difficulty on unseen skeletons.}
\begin{table}[t]
\centering
\caption{Motion-category breakdown in the diversified held-out diagnostic setting. JP is in cm and An is in degrees.}
\label{tab:diverse_unseen}
\begin{tabular}{lcc}
\toprule
Split & JP & An \\
\midrule
Zoo-Seen & 2.51 & 10.46 \\
Zoo-Rare & 3.37 & 12.46 \\
Zoo-Unseen (all) & 4.20 & 15.35 \\
\quad common motions (37) & 3.90 & 13.56 \\
\quad dynamic motions (29) & 4.58 & 17.65 \\
Obj & 4.93 & 11.14 \\
\bottomrule
\end{tabular}
\end{table}
We analyze 66 valid sequences from the six excluded species. Table~\ref{tab:diverse_unseen} shows that dynamic motions are harder than common motions (17.65 vs.\ 13.56$^{\circ}$), confirming that motion difficulty substantially affects unseen-skeleton results.

\paragraph{Reference sensitivity.}
\begin{table}[t]
\centering
\caption{Reference sensitivity in the diversified held-out diagnostic setting. Each cell reports JP/An; five-reference results are mean$\pm$std. ZS, ZR, and ZU denote Zoo-Seen, Zoo-Rare, and Zoo-Unseen.}
\label{tab:ref_sensitivity}
\resizebox{\columnwidth}{!}{
\begin{tabular}{lccc}
\toprule
Split & 5 refs. (1/5/10/15/20th) & noise std 0.05--0.40 & rest+identity \\
\midrule
ZS  & $2.44{\pm}0.04/9.36{\pm}0.64$  & $2.51/11.0$--$14.5$ & $2.51/13.9$ \\
ZR  & $3.23{\pm}0.09/11.75{\pm}0.41$ & $3.37/12.9$--$17.1$ & $3.37/17.3$ \\
ZU  & $4.10{\pm}0.06/14.39{\pm}0.58$ & $4.20/16.0$--$19.5$ & $4.20/19.4$ \\
Obj & $4.91{\pm}0.04/10.86{\pm}0.43$ & $4.93/11.9$--$18.7$ & $4.93/19.1$ \\
\bottomrule
\end{tabular}}
\end{table}
Training randomly samples the reference frame. Table~\ref{tab:ref_sensitivity} shows that reference-frame choice has little effect, with angular-error standard deviations of $0.41$--$0.64^{\circ}$. Rotation noise causes smooth degradation, whereas the degenerate rest+identity reference produces a clear drop. Joint-position error remains nearly unchanged because the reference rotation primarily affects P$\to$R.

\paragraph{Isolating Pose-to-Rotation.}
\begin{table}[t]
\centering
\caption{P$\to$R isolation on predicted positions in the diversified held-out diagnostic setting. Values are angular error in degrees.}
\label{tab:p2r_isolation}
\resizebox{\columnwidth}{!}{
\begin{tabular}{lcccc}
\toprule
Method (with reference and tracking) & ZS & ZR & ZU & Obj \\
\midrule
Analytic swing/aim & 25.2 & 26.1 & 33.0 & 34.2 \\
CCD & 26.2 & 26.7 & 33.9 & 36.8 \\
FABRIK & 25.8 & 26.7 & 33.3 & 37.5 \\
\textbf{Ours (learned P$\to$R)} & \textbf{10.5} & \textbf{12.5} & \textbf{15.4} & \textbf{11.1} \\
\bottomrule
\end{tabular}}
\end{table}
We replace learned P$\to$R with three standard IK solvers on the same predicted positions. Each receives the same reference pose--rotation pair, temporal tracking, and differentiable temporal smoothing. Table~\ref{tab:p2r_isolation} shows a consistent gap across solvers: IK recovers bone directions but cannot infer per-frame dynamic twist from video and remains sensitive to V$\to$P noise. Even with ground-truth positions, analytic IK reaches only about $14$--$15^{\circ}$, further confirming the ambiguity bottleneck.

\subsection{Efficiency Analysis}
\label{sec:exp_efficiency}

Our efficiency gain comes from eliminating both the mesh prediction stage and the analytical IK solver. For a 120-frame input sequence, V1 takes over 20 minutes (feature extraction $\sim$40\,s, mesh reconstruction $\sim$15\,min, pose estimation $\sim$20\,s, IK optimization $\sim$5\,min); our method shares the same feature-extraction cost but predicts pose and rotation in a single forward pass within $\sim$10\,s, totaling under 1 minute and corresponding to a $\sim$20$\times$ speedup. Removing the mesh avoids a heavy, error-prone reconstruction stage, while a learned rotation decoder replaces iterative per-frame IK with efficient batched computation. This speedup does not cost accuracy: averaged over Zoo-Seen/Rare/Unseen (Table~\ref{tab:exp_v1_compare}), our method achieves $10.6^{\circ}$ angle error vs.\ $18.9^{\circ}$ for V1 (GT Mesh) and $20.63^{\circ}$ (Pred Mesh).


\subsection{Qualitative Results}
\label{sec:exp_qualitative}

\noindent\textit{Qualitative figures (Figs.~\ref{fig:mocap_compare}, \ref{fig:mocap_demo}, \ref{fig:dance_demoA}, \ref{fig:dance_demoB}) are placed at the end of the paper for layout reasons. Each cell stacks three keyframes via diagonal offset and alpha fade to convey motion in a static figure; full animations are available on our project homepage.}

\paragraph{Rotation quality: V1 vs.\ V2.}
Figure~\ref{fig:mocap_compare} compares our learning-based rotation recovery (V2) against the traditional IK-based optimization used in V1. V1 frequently exhibits joint spinning and limb flipping, particularly on joints with large twist components, since per-frame IK lacks temporal context and reference priors. In contrast, V2 produces natural, temporally continuous rotations, confirming the effectiveness of reference-conditioned rotation modeling.

\paragraph{MoCap across Objaverse, Zoo, and in-the-wild.}
Figure~\ref{fig:mocap_demo} shows mocap results across three domains: Objaverse assets (row~1), Truebones Zoo (row~2), and in-the-wild Internet videos (rows~3--4), rendered from multiple viewpoints. Despite large variations in appearance, shape, and capture conditions, our method produces accurate and temporally consistent 3D motion, demonstrating that it generalizes to arbitrary subjects rather than being tied to a specific training distribution. In-the-wild examples with moving cameras in the supplemental video show stable camera-coordinate pose predictions.

\paragraph{Unified mocap and cross-skeleton retargeting.}
A distinctive feature of our framework is that a \emph{single} input video can simultaneously drive mocap and retargeting onto many different skeletons, without any skeleton-specific training. Figures~\ref{fig:dance_demoA} and~\ref{fig:dance_demoB} illustrate this on two dance clips: given the input in the center, the surrounding cells show the same motion retargeted onto a diverse set of humanoid and animal characters. The retargeted motion preserves the rhythm and semantics of the source while respecting each target skeleton's topology, enabling flexible cross-species motion transfer from a single video.

\section{Limitations}
\label{sec:limitations}

We see a few natural directions for further improvement. First, the Pose-to-Rotation decoder implicitly learns plausible per-skeleton motion priors from data, so for unnatural retargeting cases (e.g., transferring a bird's flapping motion to a dog so that it ``flies'' with its forelegs spread out) the predicted rotations tend to drift toward more typical configurations for that skeleton, even when Video-to-Pose recovers the intended pose; in the opposite direction, the human skeleton can indeed reproduce such unusual motions. Supporting more unnatural retargeting cases would mainly call for augmenting the training set with such configurations. Second, unlike fixed-topology methods, our topology-general design cannot rely on fixed anatomical or mesh priors. Severe occlusion therefore remains challenging; compatible occlusion augmentation and explicit geometry priors are promising improvements. Third, our current animal dataset contains only $\sim$1{,}000 sequences, and we expect results to improve further as this scales up.

\section{Conclusion}
\label{sec:conclusion}

We have presented the first fully end-to-end framework for arbitrary-skeleton motion capture from monocular video, in which both Video-to-Pose and Pose-to-Rotation are learnable and jointly trained. The key enabler is a single reference pose–rotation pair from the target asset, which anchors the coordinate axes of each joint's local frame and thereby converts the ill-posed P$\to$R mapping into a well-constrained conditional prediction task. Once P$\to$R is learnable, joint training lets the intermediate pose representation reshape itself to serve the final rotation objective. This is something factorized pipelines, with their non-differentiable IK handoff, cannot achieve.

On Truebones Zoo and Objaverse, our approach reduces the average rotation angle error from $\sim$17$^{\circ}$ (V1's factorized learned-V$\to$P + analytical-IK pipeline) to $\sim$10$^{\circ}$, with the largest improvements on unseen skeletons whose axis conventions rest-pose structure alone cannot disambiguate. Removing the mesh intermediate used by prior work as a video-to-joint bridge further yields $\sim$20$\times$ faster inference, since predicted-mesh errors compound through the pipeline rather than providing useful information.


\bibliographystyle{ACM-Reference-Format}
\bibliography{main}



\appendix
\clearpage
\section{Implementation Details}
\label{sec:appendix_impl}

We use a frozen DINOv2~\cite{oquab2023dinov2} ViT encoder as the visual backbone. The Video-to-Pose and Pose-to-Rotation modules are trained jointly end-to-end, with the rotation decoder using $L{=}8$ blocks and reference cross-attention in the first $L_{\text{cross}}{=}6$ layers. Sequences are processed with a length of $T{=}48$ frames and per-joint temporal attention uses a window size of $5$. The model is trained with Adam on $8{\times}$ V100 for $60$ epochs with a batch size of $8$, taking roughly one day. Loss weights are $\lambda_{\text{pos}}{=}\lambda_{\text{rot}}{=}\lambda_{\text{rot\_v}}{=}1.0$ and $\lambda_{\text{root}}{=}0.1$. The mixed-pose schedule uses $p_{\text{start}}{=}0.1$, $p_{\text{end}}{=}1.0$, and $E_{\text{warmup}}{=}30$. The maximum joint count is set to $150$ to accommodate the largest skeleton in our training data ($143$ joints); this limit is straightforward to adjust for other datasets.

\section{Warm-up Schedule Sensitivity}
\label{sec:appendix_warmup}

We study the sensitivity of the mixed-pose warm-up length $E_w$ in the end-to-end training strategy (\S\ref{sec:exp_E2Etraining}). Table~\ref{tab:exp_warmup} reports angle and position metrics for $E_w \in \{10, 20, 30, 40, 50\}$. Performance is consistent across $E_w \in [20, 50]$, indicating that the method is robust to the mixing schedule; we use $E_w{=}30$ by default.

\section{Model Depth}
\label{sec:appendix_depth}

We study the effect of model depth by jointly scaling both the Video-to-Pose and Pose-to-Rotation modules (Table~\ref{tab:exp_depth}). Increasing the depth from 6 to 8 layers improves rotation accuracy, while further increasing to 12 layers degrades performance across all splits. The 8-layer model achieves the best results ($6.54^\circ$ on Zoo-Unseen), suggesting a good balance between model capacity and optimization.

\section{Cross-Attention Depth}
\label{sec:appendix_cross}

We study the effect of reference cross-attention depth ($L_{\text{cross}}$) in the rotation decoder, while keeping the overall model depth fixed to 8 layers (Table~\ref{tab:exp_cross}). Without cross-attention ($L_{\text{cross}}{=}0$), the model fails on Zoo-Unseen ($23.49^\circ$), showing that reference conditioning is essential. Introducing cross-attention significantly improves performance, with the best results achieved at $L_{\text{cross}}{=}6$ ($6.54^\circ$ on Zoo-Unseen). Applying it in all layers ($L_{\text{cross}}{=}8$) brings no further gain and slightly degrades performance, indicating diminishing returns when over-conditioning the decoder.


\begin{table}[H]
\centering
\caption{Sensitivity to the warm-up length $E_w$ of the mixed-pose schedule. Position in cm; rotation in degrees ($\downarrow$). Best angle error per split in \textbf{bold}.}
\label{tab:exp_warmup}
\resizebox{\columnwidth}{!}{
\begin{tabular}{l|cccc|cccc|cccc|cccc}
\toprule
& \multicolumn{4}{c|}{\textbf{Zoo-Seen}}
& \multicolumn{4}{c|}{\textbf{Zoo-Rare}}
& \multicolumn{4}{c|}{\textbf{Zoo-Unseen}}
& \multicolumn{4}{c}{\textbf{Obj}} \\
\cmidrule(lr){2-5}\cmidrule(lr){6-9}\cmidrule(lr){10-13}\cmidrule(lr){14-17}
$E_w$
 & JP & JV & An & AV
 & JP & JV & An & AV
 & JP & JV & An & AV
 & JP & JV & An & AV \\
\midrule
10
 & 2.99 & 0.64 & 11.73 & 0.32
 & 3.61 & 0.69 & 15.18 & 0.40
 & 3.74 & 1.09 & 7.71 & 0.20
 & 4.38 & 1.33 & 11.46 & 0.29 \\
20
 & 2.42 & 0.56 & 11.18 & 0.31
 & 3.51 & 0.66 & 14.93 & 0.39
 & 3.32 & 1.00 & \textbf{6.48} & 0.17
 & 3.96 & 1.04 & \textbf{11.06} & 0.28 \\
30 (Ours)
 & 2.34 & 0.53 & \textbf{10.73} & 0.29
 & 2.98 & 0.61 & \textbf{14.38} & 0.37
 & 3.39 & 0.99 & 6.54 & 0.17
 & 3.84 & 1.05 & \textbf{11.06} & 0.30 \\
40
 & 2.59 & 0.53 & 11.08 & 0.30
 & 3.46 & 0.62 & 15.42 & 0.39
 & 4.23 & 1.00 & 7.10 & 0.19
 & 4.09 & 1.15 & 11.09 & 0.28 \\
50
 & 2.28 & 0.54 & 10.75 & 0.30
 & 3.01 & 0.61 & 14.42 & 0.38
 & 3.43 & 0.99 & 6.61 & 0.18
 & 4.14 & 1.10 & 11.86 & 0.29 \\
\bottomrule
\end{tabular}
}
\end{table}

\begin{table}[H]
\centering
\caption{Effect of model depth (Video-to-Pose and Pose-to-Rotation jointly scaled). Position in cm; rotation in degrees ($\downarrow$). Best angle error per split in \textbf{bold}.}
\label{tab:exp_depth}
\resizebox{\columnwidth}{!}{
\begin{tabular}{l|cccc|cccc|cccc|cccc}
\toprule
& \multicolumn{4}{c|}{\textbf{Zoo-Seen}}
& \multicolumn{4}{c|}{\textbf{Zoo-Rare}}
& \multicolumn{4}{c|}{\textbf{Zoo-Unseen}}
& \multicolumn{4}{c}{\textbf{Obj}} \\
\cmidrule(lr){2-5}\cmidrule(lr){6-9}\cmidrule(lr){10-13}\cmidrule(lr){14-17}
Depth
 & JP & JV & An & AV
 & JP & JV & An & AV
 & JP & JV & An & AV
 & JP & JV & An & AV \\
\midrule
6
 & 2.26 & 0.52 & 10.84 & 0.30 & 3.08 & 0.70 & 14.42 & 0.38 & 3.37 & 0.99 & 7.00 & 0.19 & 3.92 & 1.05 & 11.20 & 0.28 \\
8 (Ours)
 & 2.34 & 0.53 & \textbf{10.73} & 0.29 & 2.98 & 0.61 & \textbf{14.38} & 0.37 & 3.39 & 0.99 & \textbf{6.54} & 0.17 & 3.84 & 1.05 & \textbf{11.06} & 0.30 \\
12
 & 3.07 & 0.64 & 11.35 & 0.31 & 3.72 & 0.71 & 15.17 & 0.39 & 4.26 & 1.18 & 7.66 & 0.19 & 4.28 & 1.13 & 11.98 & 0.30 \\
\bottomrule
\end{tabular}
}
\end{table}

\begin{table}[H]
\centering
\caption{Effect of reference cross-attention depth ($L_{\text{cross}}$) in an 8-layer rotation decoder. Position in cm; rotation in degrees ($\downarrow$). Best angle error per split in \textbf{bold}.}
\label{tab:exp_cross}
\resizebox{\columnwidth}{!}{
\begin{tabular}{l|cccc|cccc|cccc|cccc}
\toprule
& \multicolumn{4}{c|}{\textbf{Zoo-Seen}}
& \multicolumn{4}{c|}{\textbf{Zoo-Rare}}
& \multicolumn{4}{c|}{\textbf{Zoo-Unseen}}
& \multicolumn{4}{c}{\textbf{Obj}} \\
\cmidrule(lr){2-5}\cmidrule(lr){6-9}\cmidrule(lr){10-13}\cmidrule(lr){14-17}
$L_{\text{cross}}$
 & JP & JV & An & AV
 & JP & JV & An & AV
 & JP & JV & An & AV
 & JP & JV & An & AV \\
\midrule
0 & 4.56 & 0.87 & 12.32 & 0.33 & 5.30 & 0.89 & 16.32 & 0.43 & 6.40 & 1.57 & 23.49 & 0.63 & 5.37 & 1.39 & 18.34 & 0.46 \\
2 & 2.24 & 0.53 & 10.95 & 0.30 & 2.92 & 0.60 & 14.90 & 0.38 & 3.27 & 1.02 & 7.21 & 0.19 & 3.79 & 1.07 & 11.40 & 0.29 \\
4 & 2.34 & 0.53 & 10.83 & 0.30 & 2.81 & 0.57 & 14.46 & 0.37 & 3.07 & 0.97 & 7.60 & 0.21 & 3.78 & 1.05 & 11.47 & 0.29 \\
6 & 2.34 & 0.53 & \textbf{10.73} & 0.29 & 2.98 & 0.61 & 14.38 & 0.37 & 3.39 & 0.99 & \textbf{6.54} & 0.17 & 3.84 & 1.05 & \textbf{11.06} & 0.30 \\
8 & 2.58 & 0.57 & 11.11 & 0.30 & 3.08 & 0.63 & \textbf{13.94} & 0.37 & 3.43 & 1.02 & 7.47 & 0.20 & 4.22 & 1.16 & 11.22 & 0.29 \\
\bottomrule
\end{tabular}
}
\end{table}

\end{document}